# Linear Discriminant Analysis in Credit Scoring: A Transparent Hybrid Model Approach

Md Shihab Reza[1*] Monirul Islam Mahmud[2*], Ifti Azad Abeer[3], and Nova Ahmed[4]

[1, 2, 3, 4] Design Inclusion & Access Lab (DIAL), Department of Electrical and Computer Engineering (ECE)
North South University, Dhaka, Bangladesh
[2] Department of Computer and Information Science, Fordham University, New York, USA
shihab.reza@northsouth.edu[1], mahmudislam2025@gmail.com[2], ifti.azad@northsouth.edu[3], nova.ahmed@northsouth.edu[4]

*Abstract*—The development of computing has made credit scoring approaches possible, with various machine learning (ML) and deep learning (DL) techniques becoming more and more valuable. While complex models yield more accurate predictions, their interpretability is often weakened, which is a concern for credit scoring that places importance on decision fairness. As features of the dataset are a crucial factor for the credit scoring system, we implement Linear Discriminant Analysis (LDA) as a feature reduction technique, which reduces the burden of the model's complexity. We compared 6 different machine learning models, 1 deep learning model, and a hybrid model with and without using LDA. From the result, we have found our hybrid model, 'XG-DNN,' outperformed other models with the highest accuracy of 99.45% and a 99% F1 score with LDA. Lastly, to interpret model decisions, we have applied 2 different explainable AI techniques named LIME (local) and Morris Sensitivity Analysis (global). Through this research, we showed how feature reduction techniques can be used without affecting the performance and explainability of the model, which can be very useful in resource-constrained settings to optimize the computational workload.

## I. Introduction

Credit scoring models serve as a data-driven mechanism to assess the creditworthiness of borrowers based on their financial situation. As financial institutions have to assess lots of credit applications regularly, these models provide an automatic and efficient way to reduce their workload. Researchers explored the ways to predict creditworthiness using machine learning (ML) and deep learning (DL). Numerous public datasets and online ML competitions have also contributed significantly to this domain. The most popular and largest dataset has been the Lending Club Dataset [14]. This data set has more than 2.5 million instances and 142 features representing thousands of loans made through a peer-to-peer (P2P) lending platform named Lending Club, which was once the world's largest P2P lending platform. The size and diversity of the dataset allowed researchers to explore many critical areas of credit scoring, which include evaluating and benchmarking different algorithms, applying explainable AI (XAI) techniques, and experimenting with different innovative approaches for effective and fair scoring. Works that focused on evaluating and comparing the performances of different models applied several classification techniques, validation approaches, and utilized several evaluation metrics for doing so. Teply and Polena used 10 classification techniques with 5-fold cross-validation and evaluated model performance using 6 different evaluation metrics [10]. They found logistic regression (LR), artificial neural networks (ANN), and linear discriminant analysis (LDA) to be the three best algorithms. Lyócsa et al. made a similar comparison, comparing profit scoring (PS) with traditional credit scoring (CS) in P2P lending to improve investor profitability [5]. They found that PS outperformed CS by generating higher returns (24.0% for Bondora and 15.5% for Lending Club) and greater accuracy (6.7% for Bondora and 3.1% for Lending Club). Another study that undertook the comparison of different models by Swati [11] showed that ensemble classifiers and neural networks were the most effective.

Previous research on the Lending Club dataset has explored model performance, explainability, dimensionality reduction, and feature engineering separately. However, no study has examined how combining these approaches impacts both performance and explainability. For instance, how feature engineering or dimensionality reduction affects model explainability and balances credit scoring requirements. This work addresses this gap by exploring the performance-explainability tradeoff through LDA and XAI. We applied six ML models, one deep neural network, and a novel hybrid model, comparing results with and without LDA. To enhance transparency, we utilized LIME and Morris Sensitivity Analysis.

Our analysis reveals that without LDA, XGBoost and XG-DNN are the top-performing models across accuracy, sensitivity, specificity, and balanced metrics. However, with LDA, the LDA-based XG-DNN outperforms XGBoost in accuracy, precision, recall, and F1 score. These findings highlight the value of integrating LDA with advanced ML techniques for improved performance. Additionally, LIME and Morris Sensitivity Analysis show that both top models rely on key features, such as the remaining outstanding principal, in their predictions. This study explores combining hybrid models with XAI techniques, addressing the tradeoff between performance and explainability, and opens new avenues for fair and effective credit scoring research.

## II. Literature Review

### A. Machine Learning Approaches for Credit Scoring

Different ML approaches have been emphasized by researchers in this domain. Lessmann et al., in Stefan's studies [15], assessed and updated the performance of state-of-the-art classification algorithms through a comprehensive approach of benchmarking in the context of Credit Scoring (CS), where they evaluated a range of state-of-the-art classification

---

algorithms like Decision Trees, SVM, Random Forests, and Neural Networks (NN). Several works aimed to introduce innovative ways for effective and fair prediction in CS, where Jen-Ying introduced the Reweighing algorithm to mitigate racial bias for the multi-class CS problem [9]. A novel approach called the Online Integrated Credit Scoring Model (OICSM) by integrating Gradient Boosting Decision Tree (GBDT) and NN is proposed by Zaimei Zhang et al. [12]. They compare the performance of OICSM with several baseline models. Monir El Annas et al. [3] explored the application of multi-dimensional hidden Markov models (MDHMM) in CS for P2P lending platforms and demonstrated competitive performance. Even with oversampling, MDHMM maintains promising performance, highlighting its robustness with imbalanced data. Sample imbalance was also addressed by Shih et al. by including rejected applicants in the modeling process [8]. Monir El Annas et al. also proposed a new method for semi-supervised learning, incorporating hidden Markov models (SSHMM) to tackle reject inference [4]. The Transfer Learning with Lag (TLL) algorithm is also used by Alasbahi and XiaoLin Zheng [1] on the Lending Club, German, Default, and PPDai datasets that enable knowledge transfer when the feature number changes.

### B. Approaches for Feature Selection and Explainability

Several works primarily focused on feature selection techniques or explainability of the models. Jasmina Nalić et al. used different feature section techniques named classifier feature evaluation, correlation feature evaluator, gain ratio feature evaluator, information gain feature evaluator, and relief feature evaluator in a real-life dataset of a microfinance institution [21]. M. I. Mahmud et al. [19] demonstrated how effective feature selection techniques, such as Lasso L1, enhance model accuracy and scalability in predictive systems. Shrawan also used different feature selection techniques: information gain, gain ratio, and chi-square in ML classifiers [16]. They utilized two advanced post-hoc model-agnostic explainability techniques, LIME and SHAP. Likewise, the GIRP, Anchors, ProtoDash, and SHAP+GIRP methods were implemented by Demajo et al. with the XGBoost model [2]. Branka Hadji et al. also used SHAP and LIME to interpret the model's decision using logistic regression, XGBoost, random forest, support vector machine, and NN classifier [6]. Vincenzo et al. compared five different XAI techniques, such as LIME, Anchors, SHAP, BEEF, and LORE, and explored feature importance [7]. Whereas Tanantong and Loetwiphut explored the detection of significant patterns and relationships using association rule mining, feature importance is ranked by RFECV [13]. These works mostly focused on exploring the explainability of different models and comparing different XAI techniques.

We extend the literature exploring the explainability of the models in the Lending Club dataset coupled with feature reduction techniques and show the importance of its different features based on the use of feature selection techniques.

### III. METHODOLOGY

We gathered publicly available Lending Club Loan Dataset from Kaggle. It has 2,925,492 rows and 142 features. We preprocessed the data such as cleaning data, null value handling and checking for data imbalance. Then we applied the models and LDA. Moreover, we compared the performance using different evaluation criteria. Lastly, we applied LIME and Morris Sensitivity Analysis to interpret model result. The overall workflow diagram is shown in Figure 1.

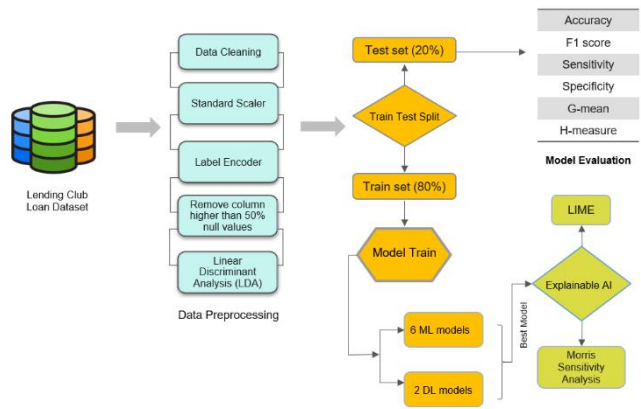

Figure 1: Overall Project Workflow

### A. Dataset Details & Preprocessing

After collecting the Lending Club dataset [14], we removed features with over 50% null values, reducing the total from 142 to 107 features. The distribution of the target variable, loan_status, was analyzed using a count plot, revealing a significant class imbalance. Loan_status comprises 10 categories. To address this imbalance, we applied the SMOTE technique, and the updated balanced distribution is shown in Figure 2.

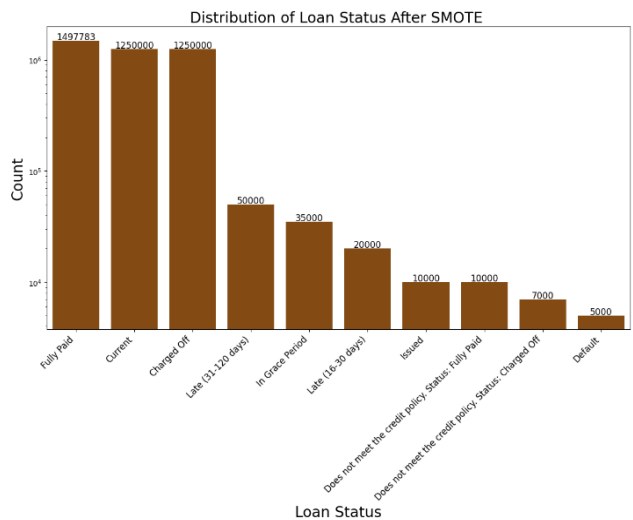

Figure 2: Distribution plot of the target variable "loan_status" after applying SMOTE.

Then, we analyzed the bar plot and histogram of categorical and numerical features to observe data distribution. Figure 3 illustrates the bar plots for the *purpose* and *emp_length* features. The *purpose* bar plot shows that the *debt_consolidation* category dominates, while the *emp_length* bar plot indicates that employees with 10+ years of experience are prioritized for loans. Moreover, in the histogram, the loan amount count is highest in the range of 9000 to 10000, and the installation count is highest in the range between 250 and 300. To ensure all features are on a similar scale between 0 and 1, we used the standard scaler technique and label encoder to convert the target variable into numerical format. Figure 4(a) and 4(b) shows the histogram of numerical features loan_amnt and installment respectively.

## B. Models Applied and Evaluation

To run different ML models, we split our dataset into a training set (80%) and a test set (20%). In this paper, we applied 8 different ML and deep learning approaches, such as XGBoost, Random Forest, Decision Tree, Naive Bayes, Logistic Regression, LDA, Deep Neural Network (DNN), and a hybrid approach named 'XG-DNN,' a mixture of XGBoost and DNN. The hybrid model 'XG-DNN' combines the strengths of two powerful algorithms, such as XGBoost and DNN.

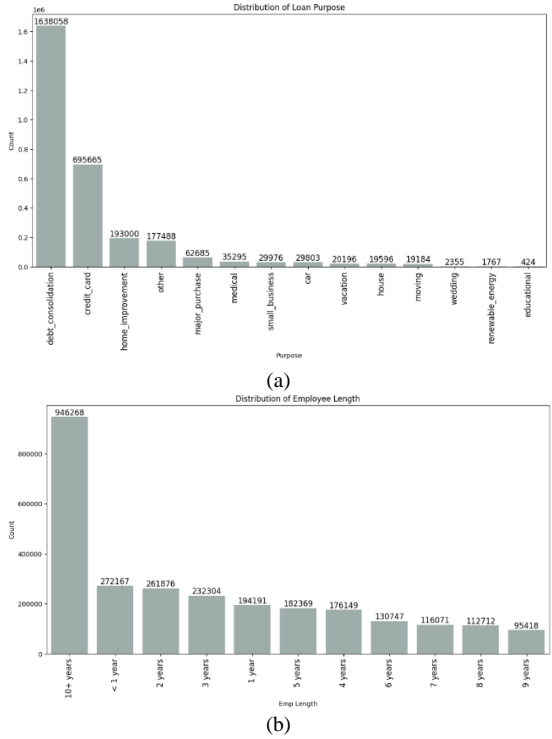

Figure 3: Bar Plot of categorical features - (a) purpose (b) emp length.

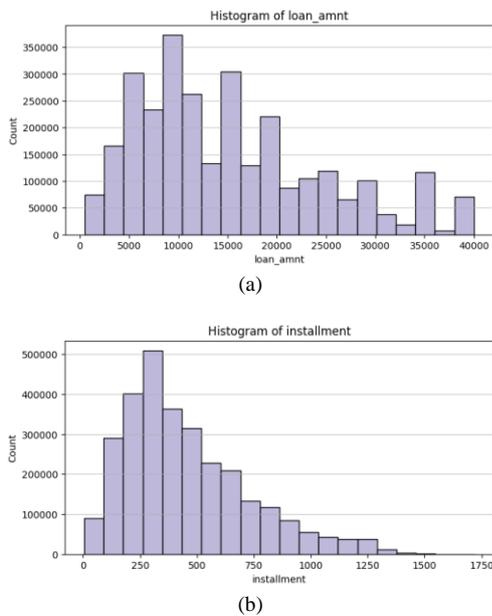

Figure 4: Histogram of numerical features (a) loan_amnt, (b) installment

XGBoost is an optimized implementation of the gradient boosting algorithm. It includes regularization techniques that prevent overfitting, ensuring the model generalizes well to unseen data, which is crucial in financial applications for transparency and trust. Conversely, DNNs are a class of DL models capable of capturing complex, non-linear relationships in data through multiple layers of interconnected neurons. They are particularly useful for financial data with complex interactions between features. By combining these models, we utilize XGBoost's ability to handle structured data and provide feature importance with DNN's capacity to learn complex patterns. This stage-wise learning helps to refine the features and capture higher-order interactions that might be missed by individual models. The hybrid model architecture is shown in Figure 5.

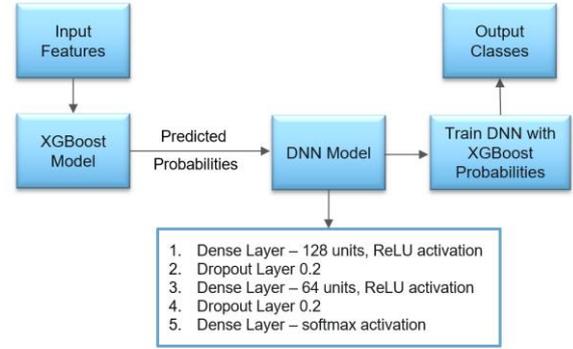

Figure 5: Architecture of hybrid XG-DNN model.

For result evaluation, we have used six state-of-the-art evaluation techniques for credit scoring: accuracy, sensitivity (true positive rate), specificity (true negative rate), G-mean, H-measure, and F1 score.

## C. Linear Discriminant Analysis (LDA)

We used LDA as a classification model and also as a feature reduction technique. While preserving the feature information as much as possible, LDA reduces the dimensionality of data in supervised classification tasks. It helps to identify linear combinations between features while changing the data from a higher-dimensional space to a lower-dimensional one. Higher-dimensional space creates overfitting, which forces a model to learn from noisy and irrelevant data, causing lower accuracy in testing and higher accuracy in training. High dimensionality can also cause higher computational complexity, data sparsity, and higher model complexity. After instantiating the LDA class, we transformed the features into the LDA space that provides a lower-dimensional data representation, which leaves us with 21 features from 107 features in our dataset. Then, we applied the XGBoost and XG-DNN hybrid models to the transformed dataset to evaluate their performance. Both models had previously performed well without using the LDA feature reduction technique.

## D. Explainability of Model

To explain the model's predictions, we used the XAI techniques named LIME (Local Interpretable Model-agnostic Explanations) and Morris Sensitivity Analysis in this work. LIME is both model-agnostic and local, meaning it can be applied to any model to explain individual predictions rather than the entire model. This provides detailed insights into the reasons behind specific outputs, making the model more transparent and understandable. Whereas Morris Sensitivity Analysis is a Global sensitivity analysis [17] that adjusts the level (discretized value) of a single input per run, named One-step-at-a-time (OAT). It was employed to identify the most influential features by calculating elementary effects to help understand the impact of variations in input parameters on the output, focusing on key parameters efficiently. By combining LIME and Morris Sensitivity Analysis, we enhanced the explainability and trustworthiness of our model both globally and locally.

## IV. RESULT ANALYSIS & DISCUSSION

### A. Models without Linear Discriminant Analysis (LDA)

After analyzing the performance of the different models using the dataset (shown in Figure 6), we concluded that XG-DNN and XGBoost are the top-performing models in terms of accuracy. They exhibit high accuracy, sensitivity, specificity, and balanced performance. XG-DNN, a hybrid model, in particular, achieved the highest accuracy among all models. Its H-measure also indicated a good balance between sensitivity and specificity, which is particularly important for CS.

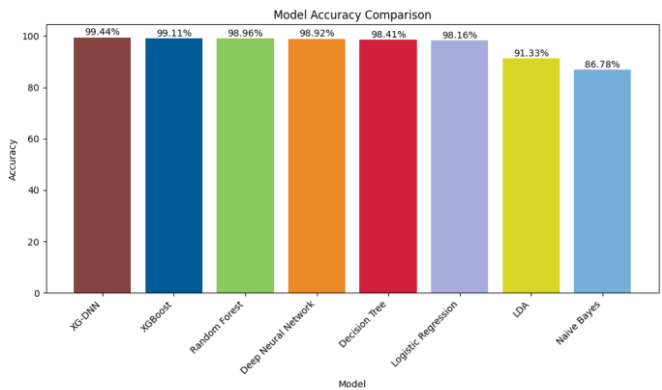

Figure 6: Comparison of accuracy of different models.

Based on accuracy, the top-performing models are XG-DNN (99.44%), XGBoost (99.11%), Random Forest (98.96%), and DNN (98.92%), while Naive Bayes had the lowest accuracy (86.78%). The LDA as a classifier model lags behind the better-performing models with an accuracy of 91.33%. Table I displays the result evaluation for the dataset across different models. Regarding sensitivity and specificity, DNN (100%) took the leading position with 100% sensitivity and specificity. In terms of G-mean evaluation metrics, the top-performing models are Deep Neural Network (100%), Logistic Regression (99.97%), Naive Bayes (99.96%), and Decision Tree (99.99%), which suggests that these models perform well in finding positive cases and excluding negative cases. Finally, regarding the F1 Score, the top-performing models are XG-DNN (99.01%) and XGBoost (99%).

### B. Best Performing Model with Linear Discriminant Analysis (LDA)

In our study, we examined how LDA affects the performance of the two best-performing models (e.g., XG-DNN and XGBoost) when we implement LDA for dimensionality reduction (results shown in Table II).

TABLE I. RESULT EVALUATION OF DIFFERENT MODELS WITHOUT LDA

| Model | Accuracy | Sensitivity | Specificity | G-mean | H-measure | F1 score |
|---|---|---|---|---|---|---|
| XG-DNN | 99.44 | 73.29 | 99.85 | 85.50 | 86.57 | 99.01 |
| XGBoost | 99.11 | 99.10 | 99.79 | 99.45 | 99.04 | 99.00 |
| DNN | 98.92 | 99.20 | 99.10 | 99.20 | 98.90 | 98.63 |
| RF | 98.96 | 98.96 | 98.89 | 98.93 | 98.87 | 98.96 |
| LR | 98.16 | 99.99 | 99.96 | 99.97 | 99.99 | 98.00 |
| NB | 86.78 | 100 | 99.93 | 99.96 | 99.99 | 91.00 |
| DT | 98.41 | 100 | 99.99 | 99.99 | 100 | 98.00 |
| LDA | 91.33 | 99.99 | 99.51 | 99.75 | 99.91 | 92.00 |

After analyzing various metrics in Table II, it is evident that the LDA-based XG-DNN hybrid model outperforms the LDA-based XGBoost model in terms of accuracy, precision, recall, and F1 score, despite its slightly lower sensitivity of 72.79%. In contrast, without LDA, we got 99.44% accuracy for XG-DNN, which is slightly lower than LDA-based XG-DNN with 99.45% accuracy. Moreover, we observed higher sensitivity, G-mean, H-measure, and F1 score of 73.29, 85.50, 86.57, and 99.01, respectively, without LDA. Whereas we can only find higher accuracy and higher specificity for XG-DNN after using LDA. On the other hand, the LDA-based XGBoost model exhibits the highest specificity of 0.99 and the highest G-mean and h-measure values of 0.99, while the LDA-based XG-DNN hybrid model shows the lowest specificity at 99.85 and the lowest G-mean value of 81.66. The F1 score further highlights the LDA-based XG-DNN hybrid model's superiority with a high score of 0.99, indicating a robust balance between precision and recall, whereas both the LDA and LDA-based XGBoost models exhibit lower F1 scores of 0.92.

TABLE II. RESULT EVALUATION OF DIFFERENT MODELS WITH LDA

| Model Name | Accuracy | Sensitivity | Specificity | G-mean | H-Measure | F1 Score |
|---|---|---|---|---|---|---|
| XGBoost | 97.32 | 99.99 | 99.98 | 99.99 | 99.99 | 92.00 |
| XG-DNN | 99.45 | 72.79 | 99.85 | 81.66 | 78.49 | 99.00 |

### C. Explainable AI (LIME & Morris Sensitivity Analysis)

We used LIME and Morris Sensitivity Analysis to analyze the underlying decision-making features influencing the XG-DNN hybrid model. By using XAI, we aimed to determine which features are crucial for the models to understand the predictions of all black box models. Figures 7 and 8 display the Lime and Morris Sensitivity Analysis explanation for the XG-DNN hybrid model.

This LIME explanation for the XG-DNN model analyzes a sample row to predict the probability of loan status, assigning class 5 ('Current'), indicating the borrower is making regular payments. The key features influencing this prediction are 'out_prncp' (44% importance), representing the remaining outstanding principal, 'total_rec_prncp' (20%), denoting the principal received to date, and 'acc_now_delinq' (9%), indicating the number of delinquent accounts. These features are crucial for predicting the loan status as 'Current.'

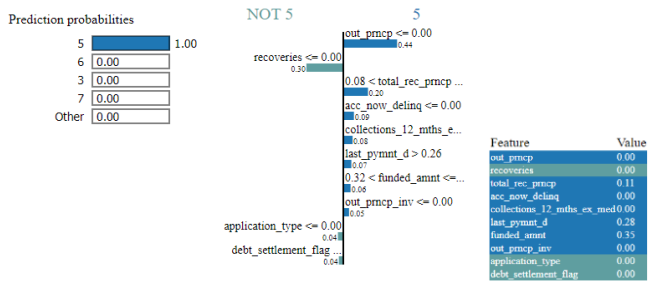

Figure 7: LIME Feature Importance for XG-DNN Model.

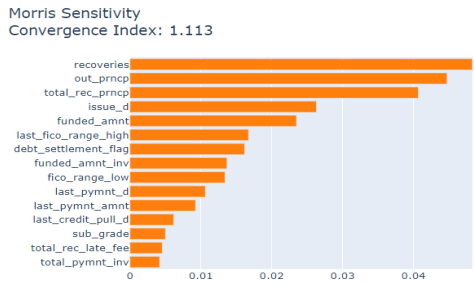

Figure 8: Morris Sensitivity Analysis for XG-DNN Model

In Morris Sensitivity Analysis for the XG-DNN model ranks features by their influence on predictions, with higher sensitivity values indicating greater impact. The most influential features are "recoveries" (0.048), "out_prncp" (0.044), and "total_rec_prncp" (0.040). Other key features include "issue_d," "funded_amnt," and "last_fico_range_high." In contrast, features like "total_pymnt_inv" (0.0041), "total_rec_late_fee" (0.0045), and "sub_grade" (0.005) show minimal impact. This analysis highlights the critical features driving the model's predictions, providing insights for feature selection and model refinement.

*D. Discussion*

Credit scoring literature has advanced rapidly in recent years, and numerous studies are being conducted for making credit scoring models effective and transparent.

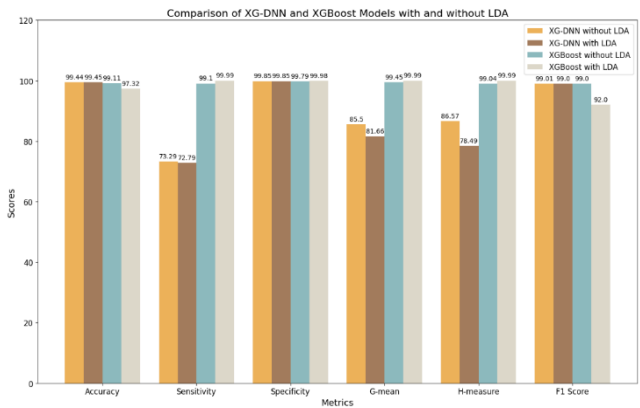

Figure 9: XG-DNN and XGBoost, with and without LDA.

Yet there are very few studies so far that have explored the impact of dimensionality reduction techniques on the performance of the models. However, these techniques are really crucial while dealing with the large, complex datasets to ensure cost-efficient data modeling. In this research, we picked the largest publicly available dataset in credit scoring to experiment with the dimensionality reduction technique, LDA, to analyze its impact on the performance of the models.

We applied 8 learning algorithms and compared them using 6 evaluation metrics in this paper to explore performance comparison across different approaches. In evaluating ML models without LDA integration, the XG-DNN hybrid model and XGBoost outperformed others, showing more balanced performance due to their ability to handle complex data relationships and mitigate overfitting. DNNs showed balanced performance with high specificity, as evidenced by their G-mean and H-measure scores. The XG-DNN hybrid achieved the highest accuracy of 99.44%, highlighting its effectiveness in combining the strengths of XGBoost and DNN. When integrating LDA with ML models, the LDA-XG-DNN hybrid showed the best results, achieving the highest accuracy and F1 score. The LDA-based XGBoost model demonstrated a favorable balance between sensitivity and specificity, reflected in a high G-mean score. Figure 9 compares the accuracy of XG-DNN hybrids and XGBoost models with and without LDA.

Despite slight trade-offs in specific metrics, the integration of LDA with XGBoost and XG-DNN models maintained strong overall classification performance, underscoring the effectiveness of LDA integration. These findings highlight the efficacy of integrating LDA with advanced ML techniques for balanced classification performance. LDA reduces the dimensionality of data and helps extract meaningful features. In large datasets, it becomes a highly crucial factor in dealing with low-resource settings, making it very useful. It enables the model to classify more accurately with fewer resources. However, it negatively impacts the transparency of the models and creates the performance explainability tradeoff. In order to address the issue, we employed both local and global XAI techniques to enhance the explainability and trustworthiness of our model. This research addresses a very important discussion avenue in credit scoring literature, namely balancing the efficiency and transparency of the models for producing non-biased predictions, where it showed that the feature reduction techniques can improve the model performance with very minimal deterioration of the classification accuracy. Moreover, it shows that the increased level of abstraction due to feature reduction and its negative impact on the transparency of the models can also be addressed by employing multiple XAI techniques. These research implications would be particularly helpful for researchers in resource-constrained settings in building effective and usable real-time prediction models with minimum computational resources.

## V. CONCLUSION & FUTURE WORK

Credit scoring literature is growing rapidly with the advancement of computing power, where learning models are becoming increasingly popular and efficient. The Lending Club dataset [14] has significantly contributed to this literature, as it is used to apply and experiment with different approaches for model performance, feature engineering, and explainability. The performance of these models is also improving significantly. However, the dilemma of ensuring the best classification performance of the model and the explainability of the classification decisions remains a challenge in this domain. This work contributes to the credit scoring literature by exploring credit scoring-specific considerations through the application of eight different machine learning and deep learning approaches. We compared their performances and analyzed how LDA, as a feature reduction technique, affects the performance of the best models, particularly on large financial datasets like the Lending Club dataset. Our study suggests that LDA is effective and helps boost the models' performance. Additionally, to understand the decision-making process of the best model, we analyzed two different XAI techniques, which are crucial for identifying the feature factors driving decisions. Moreover, this work is the initiation of an exploration with numerous future approaches. These future endeavors include, but are not limited to, applying more classification algorithms, feature selection methods such as Information Gain and Lasso L1, and dimensionality reduction techniques like PCA. We also plan to investigate various other explainable AI techniques and collect private financial data from different banks in Bangladesh to apply our study findings in the context of Bangladesh. These future explorations will help us determine the perfect balance between model performance and explainability to ensure effective and fair credit scoring predictions.